\documentclass[12pt]{article}
\usepackage{times}
\usepackage{helvet}
\usepackage{courier}
\usepackage{epstopdf}
\usepackage{color}

\usepackage{graphicx} 
\usepackage{psfrag}
\usepackage{subfigure}

\usepackage{algorithm}
\usepackage{algorithmic}

\usepackage{amsmath}
\usepackage{comment}
\usepackage{amssymb}

\newtheorem{assumption}{Assumption}
\newtheorem{theorem}{Theorem}
\newtheorem{corollary}{Corollary}
\newtheorem{lemma}{Lemma}

\newcommand{\argmin}{\mathop{\mathrm{arg\,min}{}}}
\newcommand{\argmax}{\mathop{\mathrm{arg\,max}{}}}
\newcommand{\sign}{\mathrm{sign}}
\newcommand{\soft}{\mathrm{shrink}}

\newcommand{\Expect}{\mathbb{E}}

\title{Online Classification Using a Voted RDA Method}
\author{Tianbing Xu \\ 
University of California, Irvine \\
Jianfeng Gao, Lin Xiao \\
Microsoft Research, Redmond \\
Amelia Regan \\
University of California, Irvine \\
}

\begin{document}
\maketitle

\abstract {
We propose a voted dual averaging method for online classification problems
with explicit regularization.
This method employs the update rule of the regularized dual averaging (RDA) 
method, but only on the subsequence of training examples where 
a classification error is made.
We derive a bound on the number of mistakes made by this method 
on the training set, as well as its generalization error rate.
We also introduce the concept of relative strength of regularization, and
show how it affects the mistake bound and generalization performance.
We experimented with the method using $\ell_1$-regularization
on a large-scale natural language processing task,
and obtained state-of-the-art classification performance with fairly sparse models.
}

\section{Introduction}

Driven by the Internet industry, more and more large scale machine learning 
problems are emerging and require efficient online solutions.
An example is online email spam filtering. Each time an email arrives, 
we need to decide whether it is an spam or not; after a decision is made, 
we may receive the true value feedback information from users,
and thus update the hypothesis and continue the classification 
in an online fashion. 

Online methods such as stochastic gradient descent, where
we update the weights based on each sample, would be a good choice.
The low computational cost of online methods is associated
with their slow convergence rate, which effectively introduces implicit 
regularization and is possible to prevent overfitting for very large scale
training data \cite{zhang:sgd,BoB:08}.
In practice, the optimal generalization performance of online 
algorithms often only require a small number of passes (epochs) over the 
training set.

To obtain better generalization performance, or to induce particular structure
(such as sparsity) into the solution, 
it is often desirable to add simple regularization terms to the loss function
of a learning problem. 
In the online setting, Langford et al.\ \cite{zhang:tg} proposed a 
\emph{truncated gradient} method to induce sparsity in the online gradient 
method for minimizing convex loss functions with $\ell_1$-regularization,
Duchi and Singer \cite{DuS:09} applied forward-backward splitting method 
to work with more general regularizations,
and Xiao \cite{xiao:rda} extended Nesterov's work \cite{nesterov:da} to
develop regularized dual averaging (RDA) methods. 
In the case of $\ell_1$ regularization, RDA often generates significantly
more sparse solutions than other online methods, which match the sparsity
results of batch optimization methods.
Recently, Lee and Wright \cite{leeWright:rda} show that under suitable 
conditions, RDA is able to identify the low-dimensional sparse manifold 
with high probability.

The aforementioned work provide regret analysis or convergence rate in terms 
of reducing the objective function in a convex optimization framework.
For classification problems, such an objective function is a weighted sum 
of a loss function (such as hinge or logistic loss) and a regularization
term (such as $\ell_2$ or $\ell_1$ norm).
Since the loss function is a convex surrogate for the 0-1 loss, it is often
possible to derive a classification error bound based on 
their regret bound or convergence rate.
However, this connection between regret bound and error rate
can be obfuscated by the additional regularization term.

In this paper, we propose a \emph{voted RDA} (vRDA) method for 
regularized online classification, and derive its error bounds 
(i.e., number of mistakes made on the training set),
as well as its generalization performance.
We also introduce the concept of \emph{relative strength of regularization}, 
and show how it affects the error bound and generalization performance.

The voted RDA method shares a similar structure as the 
\emph{voted perceptron} algorithm \cite{fs:margin}, 
which is the combination of the \emph{perceptron} algorithm 
\cite{rosenblatt:perceptron} and the leave-one-out  
online-to-batch conversion method \cite{helmbold:lot}.
More specifically, in the training phase, we perform the update of the RDA 
method only on the  subsequence of examples where a prediction mistake is made.
In the testing phase, we follow the deterministic leave-one-out approach, which
labels an unseen example with the majority voting of all the predictors 
generated in the training phase.
In particular, each predictor is weighted by the number of examples 
it survived to predict correctly in the training phase.

The key difference between the voted RDA method and the original RDA method
\cite{xiao:rda} is that voted RDA only updates its predictor when there is
a classification error. 
In addition to numerous advantages in terms of computational learning theory
\cite{FloydWarmuth95}, it can significantly reduce the computational cost
involved in updating the predictor.
Moreover, the scheme of update only on errors allows us to 
derive a bound on the number of classification errors that does not depend on
the total number of examples. 

Our analysis on the number of mistakes made by the algorithm
is based on the regret analysis of the RDA method \cite{xiao:rda}.
The result depends on the \emph{relative strength of regularization},
which is captured by the difference between the size of the regularization term
of an (unknown) optimal predictor, and the average size of
the online predictors generated by the voted RDA method.
In absense of the regularization term, our results matches that of
the voted perceptron algorithm (up to a small constant).
Moreover, our notion of relative strength of regularization and
error bound analysis also applies to the voted versions of other online 
algorithms, including the forward-backward splitting method
\cite{DuS:09}.

\section{Regularized online classification}
\label{sec:classification}

In this paper, we mainly consider binary classification problems.
Let $\{(x_1,y_1),\ldots,(x_m,y_m)\}$ be a set of training examples, where
each example consists of a feature vector $x_i\in \mathbb{R}^n$ 
and a label $y_i\in\{+1, -1\}$.
Our goal is to learn a classification function $f:\mathbb{R}^n\to\{+1,-1\}$
that attains a small number of classification errors. 
For simplicity, we focus on the linear predictor
\[
f(w,x) = \sign(w^T x),
\]
where $w\in\mathbb{R}^n$ is a weight vector, or \emph{predictor}.

In a batch learning setting, we find the optimal predictor~$w$ that minimizes
the following empirical risk
\[
R_\mathrm{emp}(w) = \frac{1}{m}\sum_{i=1}^m \ell(w, z_i) + \Psi(w) ,
\]
where $\ell(w, z_i)$ is a loss function at sample $z_i = (x_i, y_i)$, and $\Psi(w)$ is a regularization 
function to prevent overfitting or induce particular structure
(e.g., $\ell_1$ norm for sparsity).
If we use the 0-1 loss function   
\[
\ell(w,z) = 1\Bigl(y=f(w,x)\Bigr)
= \left\{ \begin{array}{ll}
1 & \mbox{if}~y=f(w,x) \\
0 & \mbox{otherwise} \end{array} \right.
\]
then the total loss $\sum_{i=1}^m\ell(w,z_i)$ is precisely the total number of 
classification errors made by the predictor~$w$.

However, the 0-1 loss function is non-convex and thus it is very difficult to
optimize.
In practice, we often use a surrogate convex function, such as the 
\emph{hinge loss} $\ell(w,z)=\max\{0, (1 - y w^T x) \}$,
the \emph{logistic loss} $\ell(w,z) = \log_2(1+\exp(-y w^Tx))$,
or the \emph{exponential loss} $\ell(w,z) = \exp(-y w^Tx)$.
We note that these surrogate functions are upper bounds of the 0-1 loss, 
therefore the corresponding total loss $\sum_{i=1}^m \ell(w,z)$ is
an upper bound on the total number of classification errors.

In a online classification setting, 
the training examples $\{z_1, z_2,\ldots, z_t, \ldots\}$ 
are given one at a time, and accordingly,
we generate a sequence of hypotheses $w_t$ one at a time.
At each time $t$, we make a prediction $f(w_t,x_t)$
based on the previous hypothesis $w_t$, then calculate the loss
$\ell(w_t,z_t)$  based on the true label~$y_t$. 
The next hypothesis $w_{t+1}$ is updated according to some rules,
e.g., online gradient descent \cite{zinkevich:gd}, based on the information 
available up to time~$t$.
To simplify notation in the online setting, we use a subscript to indicate
the loss function at time~$t$, i.e., we write
$\ell_t(w_t) = \ell(w_t,z_t)$ henceforth.

The performance of an online learning algorithm is often measured with the 
notion of \emph{regret}, which is the difference between the total loss
of the online algorithm $\sum_t {\ell_t(w_t)}$, and the total cost 
$\sum_t {\ell_t(w)}$ for a fixed~$w$ 
(which can only be computed from hindsight).
With an additional regularized function~$\Psi$, 
the regret with respect to~$w$, after~$T$ steps, is defined as
\begin{align*}
R_T(w) \equiv 
\sum_{t =1}^T (\ell_t(w_t) + \Psi(w_t)) - \sum_{t=1}^T (\ell_t(w) + \Psi(w) )
\end{align*}
We want the regret to be as small as possible when compared with any fixed~$w$.
In the rest of this paper, we assume that all the loss functions $\ell_t(w)$
and regularization functions~$\Psi(w)$ are convex in~$w$.

\section{The voted RDA method}
\label{sec:v-RDA}

\begin{algorithm}
\caption{The voted RDA method (training)}
\label{alg:vRDAtr}
\algsetup{indent=1.5em}
\begin{algorithmic}
\STATE \textbf{input}: training set $\{(x_1,y_1),\ldots,(x_m,y_m)\}$,\\
\STATE \hspace{1.1cm} and number of epochs $N$
\STATE{\textbf{initialize}: 
        $k\gets 1$, ~$w_1\gets0$, ~$c_1\gets0$, ~$s_0\gets0$\;}
\REPEAT
\FOR{$i=1,\ldots,m$}
	\STATE{compute prediction: $\hat{y} \gets f(w_k, x_i)$ \;}
	\IF{$\hat{y}=y_i$}
		\STATE{$c_k\gets c_k + 1$\;}
	\ELSE
        \STATE{compute subgradient $g_k\in\partial \ell_i(w_k)$\;}
		\STATE{$s_k \gets s_{k-1} + g_k$\;}
	 	\STATE{update $w_{k+1}$ according to Eq.~(\ref{eqn:rda-update})}
		\STATE{$c_{k+1} \gets 1$\;}
		\STATE{$k\gets k+1$\;}
	\ENDIF
\ENDFOR
\UNTIL {$N$ times}
\STATE \textbf{output:} number of mistakes~$M$, and a list of
\STATE \hspace{1.3cm}  predictors $\{(w_1,c_1),\ldots,(w_M,c_M)\}$
\end{algorithmic}
\end{algorithm}

The voted RDA method is described in Algorithm~\ref{alg:vRDAtr} and
Algorithm~\ref{alg:vRDAt}, for training and testing respectively.
The structure of the algorithm description is very similar to the
voted perceptron algorithm \cite{fs:margin}.
In the training phase (Algorithm~\ref{alg:vRDAtr}), 
we go through the training set~$N$ times, and
only update the predictor when it makes a classification error. 
Each predictor~$w_k$ is associated with a counter~$c_k$,
which counts the number of examples it processed correctly.
These counters are then used in the testing module (Algorithm~\ref{alg:vRDAt})
as the voting weights to generate a prediction on an unlabeled example.

The update rule used in Algorithm~\ref{alg:vRDAtr} takes the same form as
the RDA method \cite{xiao:rda}:
\begin{equation}\label{eqn:rda-update}
w_{k+1}=\argmin_w\left\{\frac{1}{k} s_k^T w +\Psi(w)
    +\frac{\beta_k}{k} h(w)\right\},
\end{equation}
where $\Psi(w)$ is the convex regularization function,
$h(w)$ is an auxiliary strongly convex function, and 
\begin{equation}\label{eqn:beta-sqrt}
\beta_k=\eta\sqrt{k}, \qquad \forall\, k\geq 1,
\end{equation}
where $\eta>0$ is a parameter that controls the learning rate.
Note that~$k$ is the number of classification mistakes,
$s_k$ is the summation of subgradients for the $k$ samples with 
classification mistakes, and $c_k$ is the counter of survival times
for the predictor~$w_k$.

\begin{algorithm}[t]
\caption{The voted RDA method (testing)}
\label{alg:vRDAt}
\begin{algorithmic}
\STATE{\textbf{given:}\,weighted predictors\,$\{(w_1,c_1),\ldots,(w_M,c_M)\}\!\!\!$}
\STATE{\textbf{input:} an unlabeled instance $x$}
\STATE{\textbf{output:} a predicted label $\hat{y}$ given by:}
\begin{equation} \label{eqn:vote}
\textstyle\hat{y} = \sign \left(\sum_{k=1}^M c_k \,f(w_k,x) \right)
\end{equation}
\end{algorithmic}
\end{algorithm}

For example, with $\ell_1$-regularization, we use
\[
\Psi(w) = \lambda \|w\|_1, \qquad h(w) = \frac{1}{2}\|w\|_2^2.
\]
In this case, the update rule~(\ref{eqn:rda-update}) has a closed-form solution
that employs the shrinkage (soft-thresholding) operator:
\begin{equation}\label{eqn:l1-rda}
w_{k+1} = - \frac{\sqrt{k}}{\eta}\, \soft\left(\frac{1}{k}{s}_k, \lambda\right).
\end{equation}
For a given vector $g$ and threshold~$\lambda$,
the shrinkage operator is defined coordinate-wise as
\[
(\soft(g,\lambda) )^{(i)} = \left\{ \begin{array}{ll}
g^{(i)} - \lambda & \mbox{if}~g^{(i)}>\lambda, \\
0 & \mbox{if}~ |g^{(i)}| \leq \lambda, \\
g^{(i)} + \lambda & \mbox{if}~g^{(i)}<-\lambda,
\end{array} \right.
\]
for $i=1,\ldots,n$.
Closed-form solutions for other regularization functions can be found,
e.g., in Duchi and Singer \cite{DuS:09} and Xiao \cite{xiao:rda}.

For large scale problems, storing the list of predictors 
$\{(w_1,c_1),\ldots,(w_M,c_M)\}$ 
and computing the majority vote in~(\ref{eqn:vote}) can be very costly.
For linear predictors (i.e., $\hat y = \sign(w^T x)$), 
we can replace the majority vote with a single 
prediction made by the weighted average predictor 
$\tilde{w}_M = (1/M)\sum_{k=1}^M c_k w_k$,
\[
\hat{y} = \sign\left( \tilde{w}_M^T x \right)
= \sign\left( \frac{1}{M}\sum_{k=1}^M c_k(w_k^T x) \right).
\]
In practice, this weighted average predictor generates very similar 
robust performance as the majority vote \cite{fs:margin}, 
and saves lots of memory and computational cost.

\section{Bound on the number of mistakes}
\label{sec:mbd}

We provide an analysis of the voted RDA method for the case $N=1$
(i.e., going through the training set once).
The analysis parallels that for the voted perceptron algorithm given
in Freund and Schapire \cite{fs:margin}.
In this section, we bound the number of mistakes made by the voted RDA method
through its regret analysis.
Then in the next section, we give its expected error rate in
an online-to-batch conversion setting.

First, we recognize that the voted RDA method is equivalent to running the
RDA method \cite{xiao:rda} on the subsequence of training examples
where a classification mistake is made.
Let $M$ the number of prediction mistakes made by the algorithm after
processing the~$m$ training examples, and $i(k)$ denote the index of
the example on which the $k$-th mistake was made (by~$w_k$).
The regret of the algorithm, with respect to a fixed vector~$w$, 
is defined only on the examples with prediction errors:
\begin{equation}\label{eqn:voted-regret}
R_M(w) = \sum_{k=1}^M \bigl(\ell_{i(k)}(w_k)+\Psi(w_k)\bigr)
       - \sum_{k=1}^M \bigl(\ell_{i(k)}(w)+\Psi(w)\bigr) .
\end{equation}

According to Theorem~1 of Xiao \cite{xiao:rda},
the RDA method~(\ref{eqn:rda-update}) has the following regret bound:
\[
R_M(w) \leq \beta_M h(w) + \frac{G^2}{2}\sum_{k=1}^{M} \frac{1}{\beta_k},
\]
where $G$ is an upper bound on the norm of the subgradients, i.e., 
$\|g_k\|_2\leq G$ for all~$k=1,\ldots,M$.
For simplicity of presentation, we restrict to the case of 
$h(w)=(1/2)\|w\|_2^2$ in this paper.
If we choose $\beta_k$ as in~(\ref{eqn:beta-sqrt}),
then, by Corollary~2 of Xiao \cite{xiao:rda},
\[
R_M(w) \leq \left( \frac{\eta}{2}\|w\|_2^2+ \frac{G^2}{\eta} \right) \sqrt{M}.
\]
This bound is minimized by setting
$\eta=\sqrt{2}G/\|w\|_2$,
which results in
\begin{equation}\label{eqn:regret-sqrt}
R_M(w) \leq \sqrt{2} G \|w\|_2 \sqrt{M}.
\end{equation}

To bound the number of mistakes~$M$, we use the fact that the loss
functions $\ell_i(w)$ are surrogate (upper bounds) for the 0-1 loss. 
Therefore,
\[
M \leq \sum_{k=1}^M \ell_{i(k)} (w_k).
\]
Combining the above inequality with the definition of regret 
in~(\ref{eqn:voted-regret}) and the regret bound~(\ref{eqn:regret-sqrt}), 
we have
\begin{equation}\label{eqn:mbd-hinge}
M \leq \sum_{k=1}^M \ell_{i(k)}(w) +M \lambda\Delta(w)
+ \sqrt{2} G \|w\|_2 \sqrt{M}.
\end{equation}
where $\Delta(w)$ is the \emph{relative strength of regularization},
defined as
\begin{equation}\label{eqn:Delta}
\Delta(w) = \Psi(w) - \frac{1}{M} \sum_{k=1}^M \Psi(w_k).
\end{equation}
We can also further relax the bound by replacing $\Delta(w)$ with
$\bar\Delta(w)$, defined as
\[
\bar\Delta(w) = \Psi(w) - \Psi(\bar{w}_M), 
\]
where $\bar{w}_M = \frac{1}{M} \sum_{k=1}^M w_k$ is the (unweighted) average 
of the predictors generated by the algorithm.
Note that by convexity of $\Psi$,
we have $\Delta(w) \leq \bar\Delta(w)$.

\subsection{Analysis for the separable case}
Our analysis for the separable case is based on the hinge loss
$\ell_i(w) = \max\{0,\, 1-y_i(w^T x_i) \}$.

\begin{assumption} \label{asp:separable}
There exists a vector $u$ such that $y_i(u^T x_i) \geq 1$
for all $i=1,\ldots,m$.
\end{assumption}

This is the standard \emph{separability with large margin} assumption.
Under this assumption, we have
\[
\sum_{k=1}^M \ell_{i(k)}(u) = \sum_{k=1}^M \max\{0,1-y_{i(k)}(u^T x_{i(k)})\}=0
\]
for any $M>0$ and any subsequence $\{i(k)\}_{i=1}^M$.
The margin of separability is defined as $\gamma = 1/\|u\|_2$.
For convenience, we also let 
$$R=\max_{i=1,\ldots,m} \|x_i\|_2 \,.$$
Then we can set $G=R$ since for hinge loss, $- y_i x_i$ is the subgradient
of $\ell_i(w)$, and we have
$\|-y_i x_i\|_2 = \|x_i\|_2 \leq R$ for $i=1,\ldots,m$.
We have the following results under Assumption~\ref{asp:separable}:

\begin{itemize}
\item

If $\lambda=0$ (the case without regularization), then
$M \leq \sqrt{2} G\|u\|_2 \sqrt{M}$, which implies
\[
M \leq 2 G^2 \|u\|_2^2 = \\
2\left(\frac{R}{\gamma}\right)^2.
\]
This is very similar to the mistake bound for the voted perceptron
\cite{fs:margin}, with an extra factor of two.
Note that this bound is independent of the dimension~$n$
and the number of examples~$m$.
It also holds for $N>1$ (multiple passes over the data).
\item

If $\lambda>0$, the mistake bound also depends on $\Delta(u)$, which is
the difference between $\Psi(u)$ and the \emph{unweighted} average of 
$\Psi(w_1),\ldots,\Psi(w_M)$.
More specifically,
\begin{equation} \label{eqn:M-Delta}
M \leq  M\lambda \Delta(u) + \sqrt{2} R \|u\|_2 \sqrt{M}.
\end{equation}
Note that $\Psi(w_1),\ldots,\Psi(w_M)$
tend to be small for large values of~$\lambda$ (more regularization),
and tend to be large for small values of~$\lambda$ (less regularization).
We discuss two scenarios:

\emph{The under-regularization case:} $\Delta(u)<0$.
This happens if the regularization parameter~$\lambda$ is chosen too small,
and the generated vectors $w_1,\ldots,w_M$ on average has a larger value
of $\Psi$ than $\Psi(u)$.
In this case, we have
\[
M \leq 2 \left(\frac{1}{1+\lambda |\Delta(u)|}\right)^2
  \left(\frac{R}{\gamma}\right)^2.
\]
So we have a smaller mistake bound than the case of ``perfect'' regularization
(when $\Delta(u)=0$).
This effect may be related to over-fitting on the training set.

\emph{The over-regularization case:} $\Delta(u)>0$.
This happens if the regularization parameter~$\lambda$ is chosen too large,
and the generated vectors $w_1,\ldots,w_M$ on average has a smaller $\Psi$
value than $\Psi(u)$.
If in addition $\lambda|\Delta(u)|<1$, then we have
\[
M \leq 2 \left(\frac{1}{1-\lambda |\Delta(u)|}\right)^2
  \left(\frac{R}{\gamma}\right)^2,
\]
which can be much larger than the case of ``perfect'' regularization
(meaning $\Delta(u)=0$).
If $\lambda\Delta(u)\geq 1$, then the inequality~(\ref{eqn:M-Delta})
holds trivially and does not give any meaningful mistake bound.
\end{itemize}

\subsection{Analysis for the inseparable case}
We start with the inequality~(\ref{eqn:mbd-hinge}).
To simplify notation, let $L(u)$ denote the total loss of an
arbitrary vector~$u$ over the subsequence $i(k)$, $k=1,\ldots,M$, i.e.,
\begin{equation}\label{eqn:Lu}
L(u) = \sum_{k=1}^M {\ell_{i(k)}(u)}.
\end{equation}
Then we have
\begin{equation}\label{eqn:Mbd-L}
M \leq L(u) + M\lambda\Delta(u) + \sqrt{2} R \|u\|_2 \sqrt{M}.
\end{equation}
Our analysis is similar to the error analysis for the perceptron in
\cite{ss:online}.
which relies on the following simple lemma:
\begin{lemma}\label{lem:quadratic}
Given $a,b,c>0$, the inequality $ax-b\sqrt{x}-c\leq 0$ implies
\[
x
~\leq~\frac{c}{a} + \left(\frac{b}{a}\right)^2 + \frac{b}{a}\sqrt{\frac{c}{a}}
~\leq~ \left( \sqrt{\frac{c}{a}} + \frac{b}{a} \right)^2 .
\]
\end{lemma}
Here are the case-by-case analysis:
\begin{itemize}
\item

If $\lambda=0$ (the case without regularization), we have
\[
M ~\leq~ L(u) + \sqrt{2} R \|u\|_2 \sqrt{M},
\]
which results in 
\begin{align*}
M ~\leq~ \left( \sqrt{L(u)} + \sqrt{2}R\|u\|_2\right)^2.
\end{align*}
Note that this bound only makes sense if the total loss $L(u)$ is not too large.
\item

If $\lambda>0$, the mistake bound depends on $\Delta(u)$,
the relative strength of regularization.

\emph{The under-regularization case:} $\Delta(u)<0$.
we have
\[
M~\leq~ \left( \sqrt{\frac{L(u)}{1+\lambda|\Delta(u)|}}
+ \frac{\sqrt{2}R\|u\|_2}{1+\lambda|\Delta(u)|}\right)^2.
\]

\emph{The over-regularization case:} $\Delta(u)>0$.
If $\lambda |\Delta(u)|<1$,
we have
\[
M~\leq~ \left( \sqrt{\frac{L(u)}{1-\lambda|\Delta(u)|}}
+ \frac{\sqrt{2}R\|u\|_2}{1-\lambda|\Delta(u)|}\right)^2.
\]
Again, if $\lambda\Delta(u)\geq 1$, the inequality~(\ref{eqn:Mbd-L})
holds trivially and does not lead to any meaningful bound.
\end{itemize}

\begin{theorem}\label{thm:training}
Let $\{(x_1,y_1), \ldots, (x_m,y_m)\}$ be a sequence of labeled examples 
with $\|x_i\|_2 \leq R$.
Suppose the voted RDA method (Algorithm~\ref{alg:vRDAtr}) makes~$M$
prediction errors on the subsequence $i(1),\ldots,i(M)$, and generates 
a sequence of predictors~$w_1,\ldots,w_M$.
For any vector $u$, let $L(u)$ be the total loss defined in~(\ref{eqn:Lu}),
and $\Delta(u)$ be the relative strength of regularization defined 
in~(\ref{eqn:Delta}).
If $\lambda \Delta(u)<1$, 
then the number of mistakes $M$ is bounded by
\[
M~\leq~ \left( \sqrt{\frac{L(u)}{1-\lambda\Delta(u)}}
+ \frac{\sqrt{2}R\|u\|_2}{1-\lambda\Delta(u)}\right)^2.
\]
In particular, if the training set satisfies Assumption~\ref{asp:separable},
then we have
\[
M ~\leq~ 2 \left(\frac{1}{1-\lambda \Delta(u)}\right)^2
  \left(\frac{R}{\gamma}\right)^2,
\]
where $\gamma = 1/\|u\|_2$ is the separation margin.
\end{theorem}

The above theorem is stated in the context of using the hinge loss. 
However, the analysis for the inseparable case holds for other convex surrogate 
functions as well, including the hinge loss, logistic loss and exponential loss.
We only need to replace~$R$ with a constant~$G$, which satisfies
$G\geq\|g_k\|_2$ for all $k=1,\ldots,M$.

For a strongly convex regularizer such as $\Psi(w)=(\lambda/2)\|w\|_2^2$, 
the regret bound is on the order of $\log{M}$ \cite{xiao:rda}.
Thus, for any hypothesis $u$,
the training error bound can be derived from
\[
M (1 - \lambda \Delta(u)) \leq G \|u\|_2 \log{M} + L(u).
\]
Online SVM is a special case following the above bound with hinge loss 
and $\ell_2$ regularizer.

\section{Online-to-batch conversion}
\label{sec:online2batch}
The training part of the voted RDA method (Algorithm~\ref{alg:vRDAtr})
is an online algorithm, which makes a small number of mistakes when
presented with examples one by one (see the analysis in Section~\ref{sec:mbd}).
In a batch setting, we can use this algorithm to process the training data
one by one (possibly going through the data multiple times),
and then generate a hypothesis which will be evaluated on a separate test set.

Following Freund and Schapire \cite{fs:margin}, 
we use the deterministic leave-one-out method
for converting an online learning algorithm into a batch learning algorithm.
Here we give a brief description.
Suppose we have $m$ training examples and an unlabeled instance, all
generated i.i.d.\ at random.
Then, for each~$r\in\{0,m\}$, we run the online algorithm on a sequence
of~$r+1$ examples consisting of the first~$r$ examples in the training set
and the last one being the unlabeled instance.
This produces $m+1$ predictions for the unlabeled instance, and we take
the majority vote of these predictions.

It is straightforward to see that the testing module of the voted
RDA method (Algorithm~\ref{alg:vRDAt}) outputs exactly such a
majority vote, hence the name ``voted RDA.''
Our result is a direct corollary of a theorem from Freund and Schapire
\cite{fs:margin},
which is a result of the theory developed in Helmbold and Warmuth 
\cite{helmbold:lot}.
\begin{theorem}\cite{fs:margin} 
Assume all examples $\{(x_i,y_i)\}_{i\geq 1}$ are generated i.i.d.\ at random.
Let $E$ be the expected number of mistakes that an online algorithm makes
on a randomly generated sequence of $m+1$ examples.
Then given~$m$ random training examples, the expected probability that the
deterministic leave-one-out conversion of this online algorithm makes a mistake
on a randomly generated test instance is at most $2E/(m+1)$.
\end{theorem}

\begin{corollary}
Assume all examples are generated i.i.d.\ at random.
Suppose that we run Algorithm~\ref{alg:vRDAtr} on
a sequence of examples $\{(x_1,y_1),\ldots,(x_{m+1},y_{m+1})\}$
and $M$ mistakes occur on examples with indices $i(1),\ldots,i(M)$.
Let $\Delta(u)$ and $L(u)$ be defined as in~(\ref{eqn:Delta})
and~(\ref{eqn:Lu}), respectively.

Now suppose we run Algorithm~\ref{alg:vRDAtr} on~$m$ examples
$\{(x_1,y_1),\ldots,(x_{m},y_{m})\}$ for a single epoch.
Then the probability that Algorithm~\ref{alg:vRDAt} does not predict 
$y_{m+1}$ on the test instance $x_{m+1}$ is at most
\[
\frac{2}{m\!+\!1}\, \Expect\! \left[ \inf_{u:\,1-\lambda\Delta(u)>0}
\!\left( \sqrt{\frac{L(u)}{1\!-\!\lambda\Delta(u)}}
+ \frac{\sqrt{2}R\|u\|_2}{1\!-\!\lambda\Delta(u)}\right)^{\!\!2} \right].
\]
(The above expectation $\Expect[\cdot]$
is over the choice of all $m+1$ random examples.)
\end{corollary}

\begin{table}[t]
\centering
\caption{Comparing performance of different algorithms}
\label{tab:fscore}
\begin{center}
\renewcommand{\tabcolsep}{0.8ex}
\begin{tabular} {|p{3cm}|c|c|c|c|}\hline
Algorithms & Precision & Recall & F-Score & NNZ \\ \hline
Baseline & 0.8983 & 0.8990 & 0.8986 & N.A. \\\hline
Perceptron & 0.9191 & 0.9143 & \textbf{0.9164} & \textbf{939 K} \\\hline
TG (hinge) & 0.9198 & 0.9127 & \textbf{0.9172} & \textbf{775 K} \\\hline
TG (log)   & 0.9190 & 0.9139 & \textbf{0.9165} & \textbf{485 K} \\\hline
vRDA (hinge) & 0.9211 & 0.9150 & \textbf{0.9175} & \textbf{932 K} \\\hline
vRDA (log)   & 0.9204 & 0.9144 & \textbf{0.9174} & \textbf{173 K} \\\hline
\end{tabular}
\end{center}
\end{table}

\begin{figure} [t]\centering
\subfigure [Hinge loss] {
\centering
\includegraphics[width=0.8\linewidth]{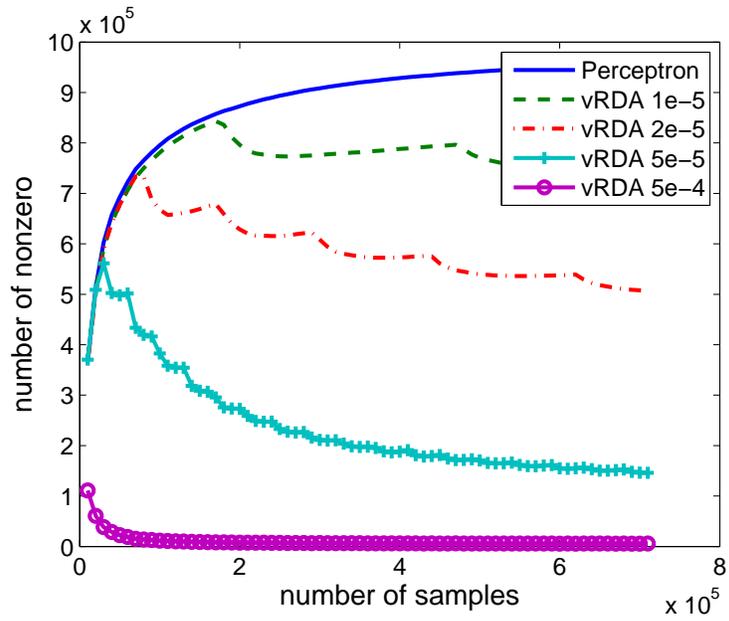}
}
\subfigure[Log loss] {
\centering
\includegraphics[width=0.8\linewidth]{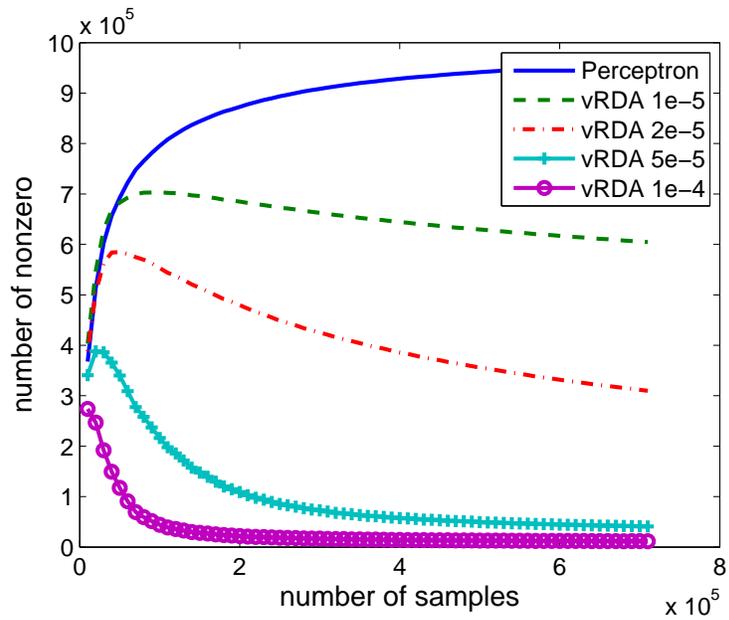}
}
\caption{Different sparse feature structure by different regularization $\lambda$ for vRDA with hinge and log losses.
The x axis is the number of samples, and the y axis shows the NNZ.
}
\label{fig:sparsity}
\end{figure}

\begin{figure}[t] 
\begin{center}
\includegraphics [width= 0.8 \linewidth]{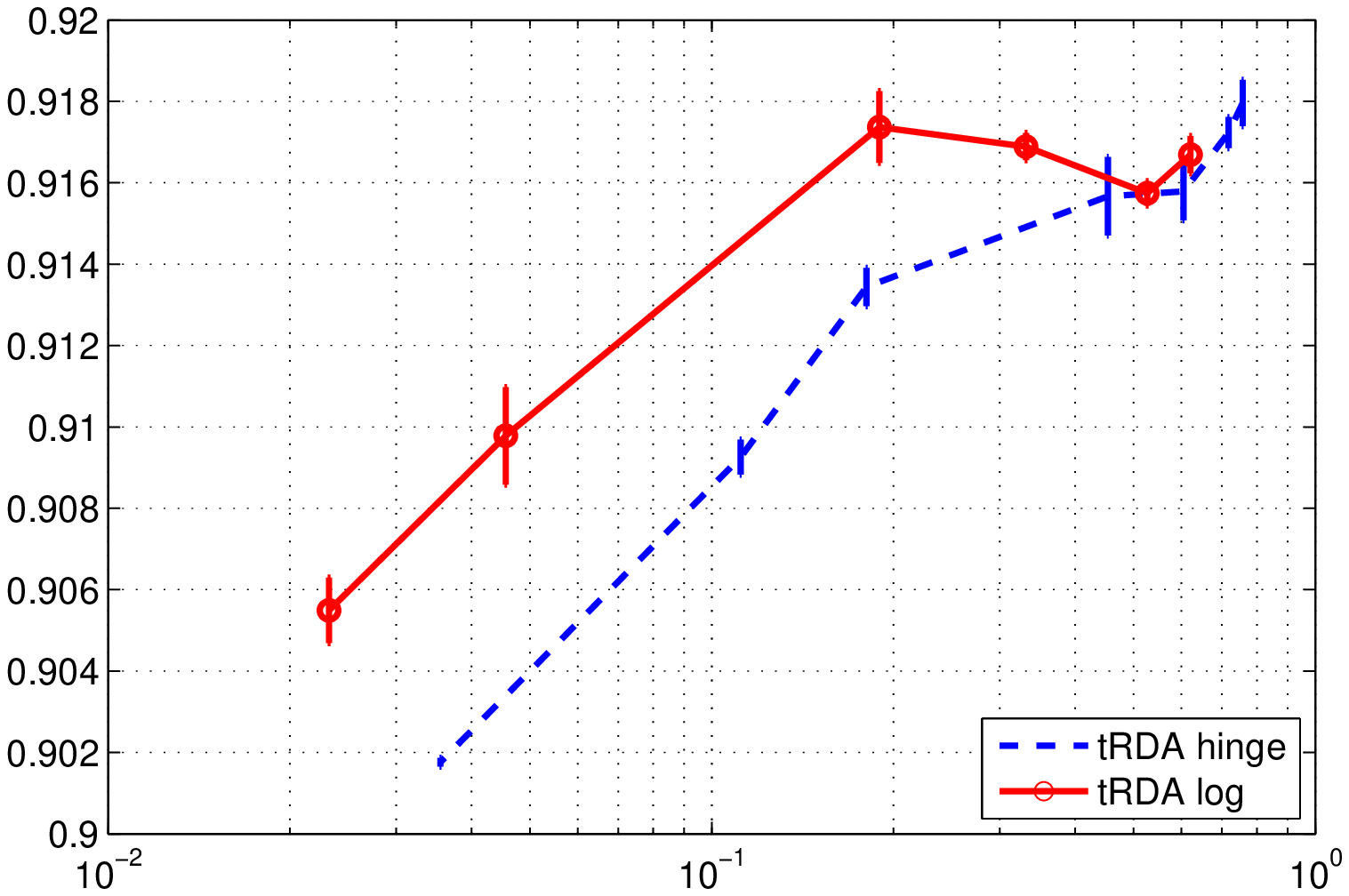}
\end{center}
\caption{Trade off between the model sparsity and classification accuracy for vRDA with hinge and log losses.
The x axis is the ratio of number of nonzero to the overall 1.2 M features; y axis is the F-Score.
}
\label{fig:FScoreSparsity}
\end{figure}

\section{Experiments on parse reranking}
\label{sec:exp}
\emph{Parse reranking} has been widely used as a test bed when adapting 
machine learning algorithms to natural language processing (NLP) tasks;
see, e.g., Collins \cite{collins:rerank}, 
Charniak and Johnson \cite{johnson:rerank}, 
Gao et al.\ \cite{gao:rerank} and Andrew and Gao \cite{andrew:rerank}.
Here, we briefly describe parse reranking as an online classification problem,
following Collins \cite{collins:rerank}.
At each time $t$, we have a sentence $s_t$ from collection of sentences $S$.
A NLP procedure is used to generate a set of candidate parses ($H_t$) 
for the sentence, and introduce a feature mapping 
$\phi(s,h): S \times H \to \mathcal{R}^n $ from the sentence and candidate
parse to a~$n$-dimensional feature vector. 
For each~$s_t$, we rank the different candidate parses based on the
linear score with a weight vector~$w$, 
and select the best parse as the one with the largest score, i.e., 
\begin{equation}\label{eqn:select-tree}
\hat{h}_t = \argmax_{h \in H_t} ~w^T \phi(s_t,h) .
\end{equation}
In the training data, we already know the oracle parse 
$h_t^*$ for $s_t$. 
If the best parse selected based on~(\ref{eqn:select-tree}) is the same 
as~$h_t^*$, we have a correct classification;
otherwise, we have a wrong classification and need to update the 
predictor~$w$.

To fit into the binary classification framework, we need to identify the 
best candidate parse other than $h_t^*$, i.e., let
\[
\tilde{h}_t = \argmax_{h \in H_t \setminus \{h_t^*\} } ~w^T \phi(s_t,h) .
\]
Then we define the feature vector for each sentence as
\[
x_t = \phi(s_t,h_t^*) - \phi(s_t, \tilde{h}_t).
\]
Therefore, if there is a classification error ($\hat{h}_t\neq h_t^*$),
we have $\tilde{h}_t=\hat{h}_t$ and $w_t^T x_t< 0$. 
Otherwise, if the classification is correct, we have 
$\tilde{h}_t\neq \hat{h}_t=h_t^*$ and $w^T x_t \geq 0$.
In summary, the binary classifier is defined as
\[
f(w,x_t) = \left\{ \begin{array}{ll}
+1 & \mbox{if}~w^T x_t \geq 0, \\
-1 & \mbox{if}~w^T x_t < 0.
\end{array} \right.
\]
Note that $w^T x_t >0$ gives the notion of a positive margin
when the classification is correct.

With the above definitions, all training examples has ``label'' $y_t = +1$.
Correspondingly, when there is a classification error (i.e., $w^T x_t<0$),
the hinge loss is
\begin{align*}
\ell_t(w) &= \max\{0,\,1 - y_t(w^T x_t)\} 
		  = \max\{0, \, 1 - w^T x_t \}
\end{align*}
Similarly, the log loss is $\ell_t(w) = \log(1 + \exp(- w^T x_t) )$.


We follow the experimental paradigm of parse reranking outlined in 
Charniak and Johnson \cite{johnson:rerank}.
We used the same generative baseline model for generating candidate parsers, 
and nearly the same feature set, which includes the log probability of a parse 
according to the baseline model and 1,219,272 additional features. We trained the 
predictor on Sections 2-19 of the Penn Treebank \cite{PennTreebank:93}, 
used Section 20-21 to optimize training parameters, 
including the regularization weight $\lambda$ and the learning rate $\eta$,
and then evaluated the predictors on Section 22. The training set contains 36K sentences, 
while the development set and the test set have 4K and 1.7K, respectively. 
Performance of parsing reranking is measured with the PARSEVAL metric, i.e., F-Score over labelled brackets.
For each epoch, we have the F-Score based on the corresponding
weights learned from these samples.
We use the weighted average of all the predictors generated by the algorithm
as the final predictor for testing.


\begin{figure} [t] \centering
\subfigure [Hinge Loss] {
\includegraphics[width = 0.8 \linewidth]{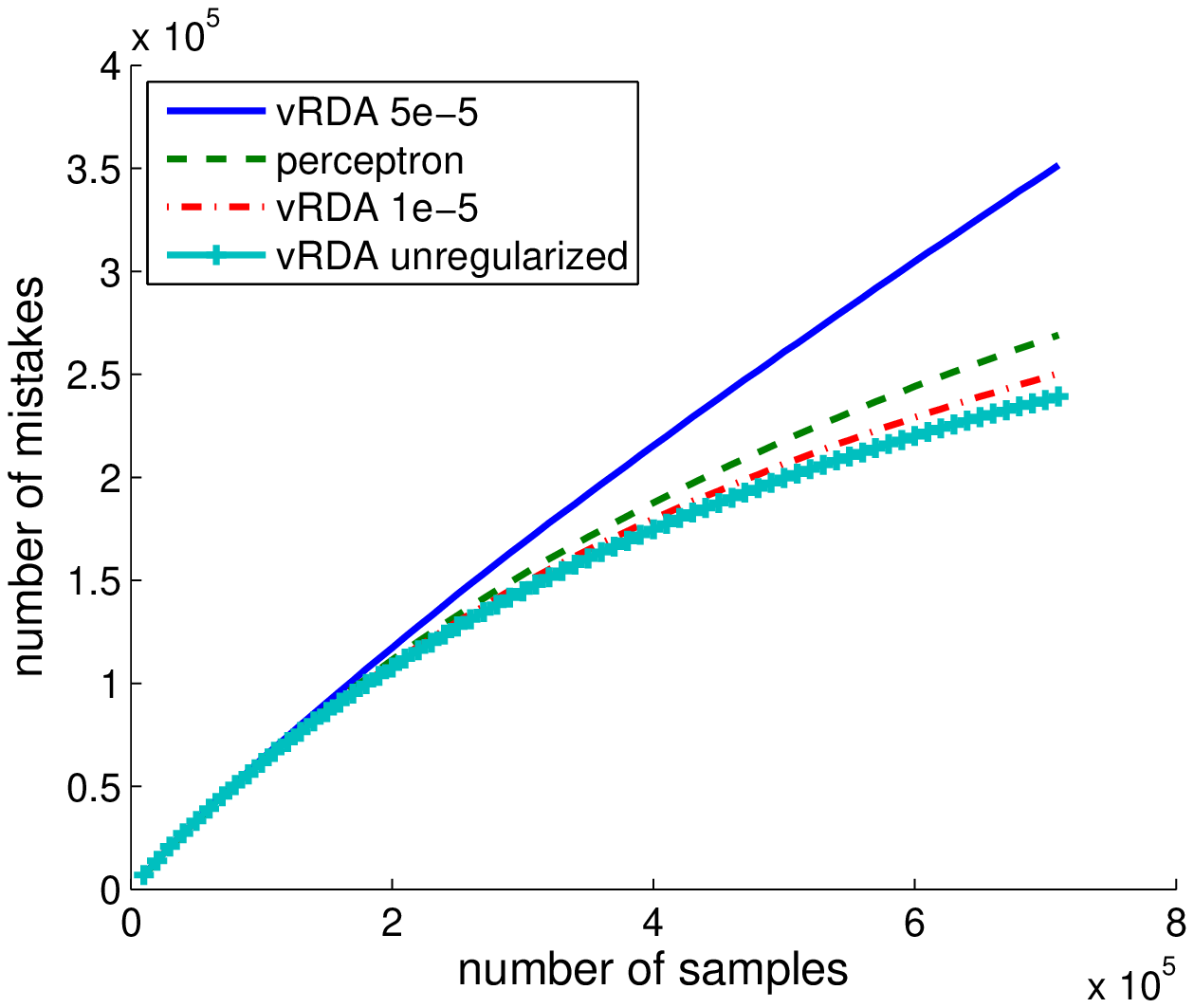}
}
\subfigure [Log Loss] {
\includegraphics[width = 0.8 \linewidth]{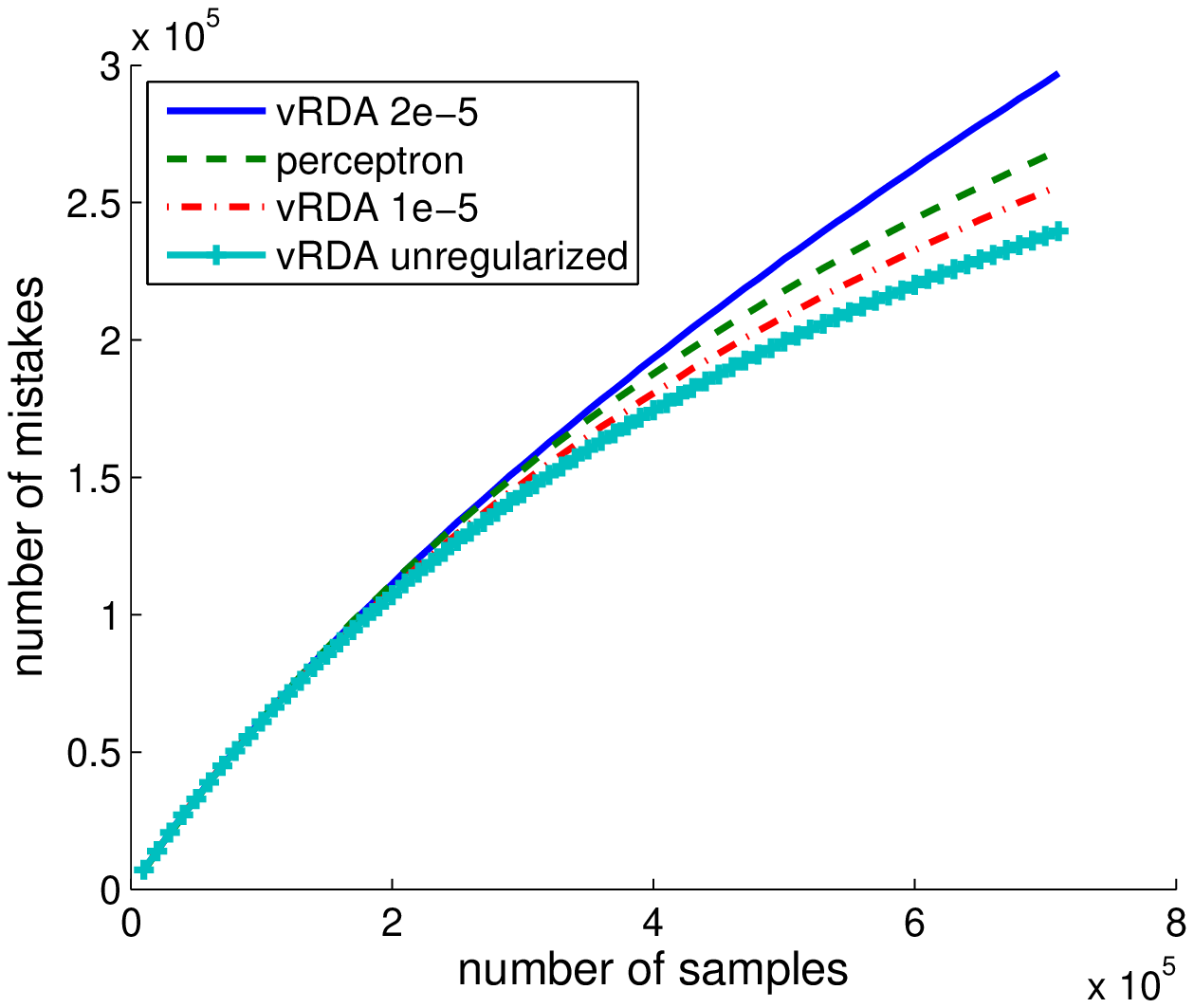}
}
\caption{Training Errors comparisons with different regularized models and \emph{voted} Perceptron 
w.r.t number of samples. The x axis is the number of samples, overall 20 epoches of data 
with 710 K samples; y axis is the number of classification mistakes.}
\label{fig:errors}
\end{figure}

\begin{figure} [t] 
\centering
\includegraphics[width = 0.8 \linewidth]{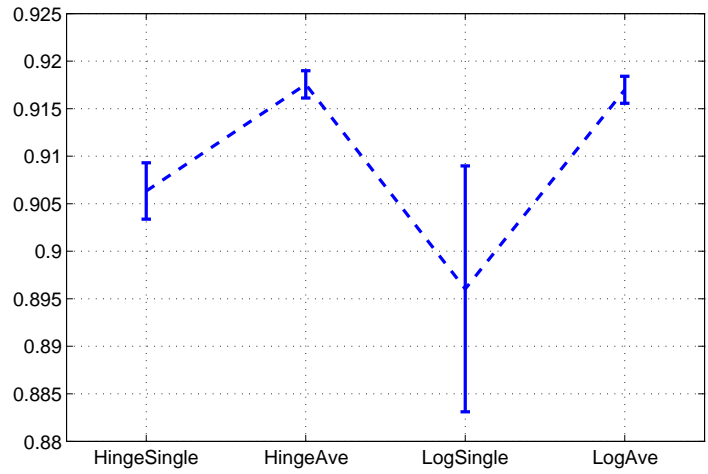}
\caption{Performance comparisons of single and average predctions for vRDA.}
\label{fig:aveComp}
\end{figure}

\begin{figure} [t] 
\centering
\includegraphics[width = 0.8 \linewidth]{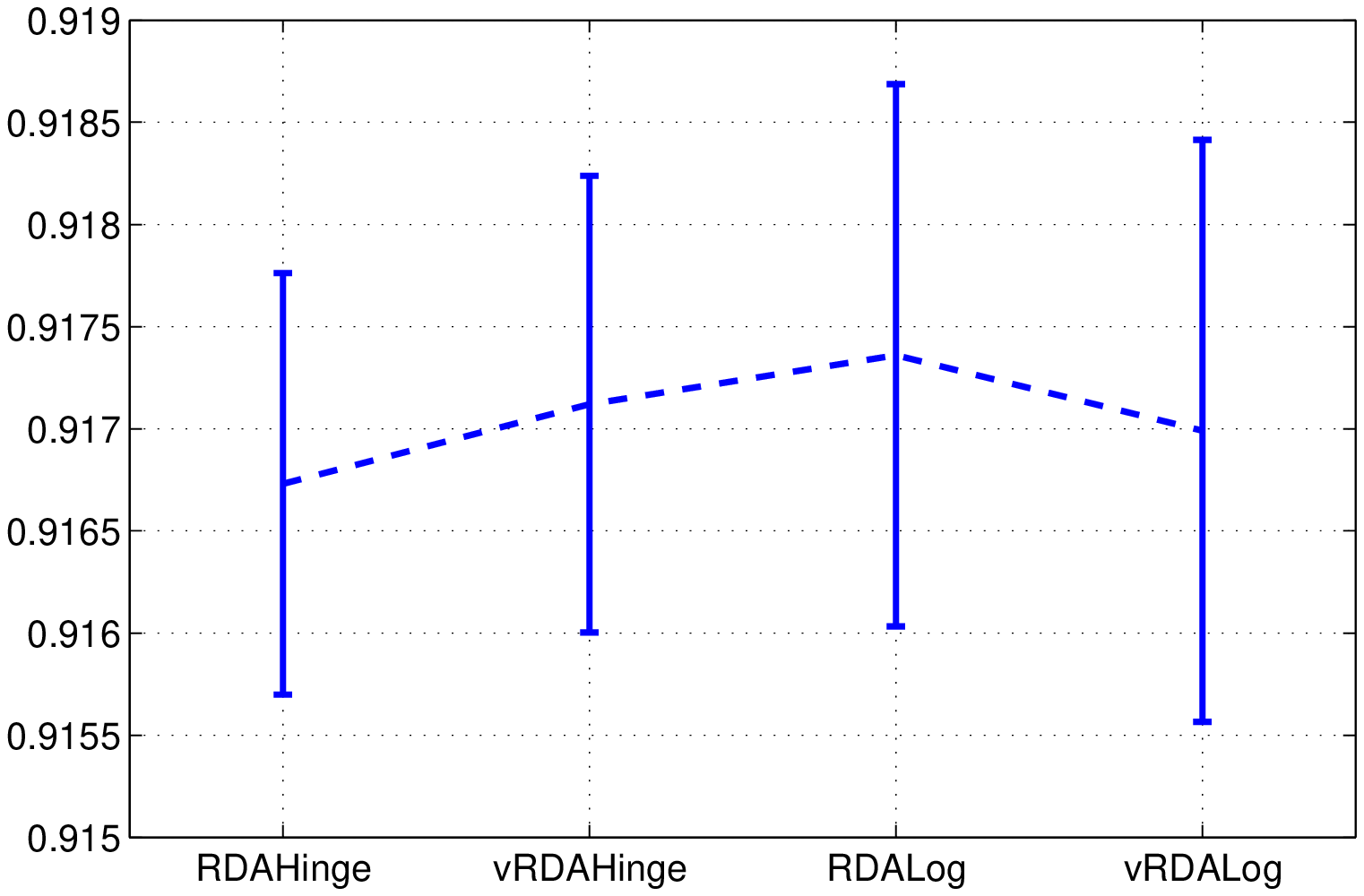}
\caption{Performance comparisons of RDA and vRDA.}
\label{fig:rdavrda}
\end{figure}

\textbf {Comparison with Perceptron and TG} 


Our main results are summarized in Tables~\ref{tab:fscore}, the F-Score and NNZ are 
averaged over the results of 20 epoches of online classification.
The baseline results are obtained by the parser in Charniak \cite{charniak:rerank}.
The implementation of perceptron follows the averaged perceptron algorithm \cite{collins:vp}. 
For voted RDA, we report results of the predictors trained using the parameter settings tuned on the development set. 
We used $\eta=0.05$ and $\lambda=1e-5$ for hinge loss, and $\eta=1000$ and $\lambda=1e-4$ for log loss. 
Results show that compared to perceptron, voted RDA achieves similar or better F-Scores with more sparse weight vectors. 
For example, using log loss we are able to achieve an F-score of 0.9174 with only 14\% of features. 
TG is the truncated gradient method {\cite{zhang:tg}}; our vRDA is a better choice than TG in term of the 
classificaiton performance and sparsity especially for log loss.


\textbf {Sparsity and Performance Trade Off} 

Since its ability to learn a sparse structured weight vector is an important advantage of voted RDA, 
we exam in detail how the number of non-zero weights changes during the course of training in 
Figure \ref{fig:sparsity}.
In vRDA, the regularization parameter $\lambda$ controls the model sparsity.
For a stronger $\ell_1$ regularizer with large values of $\lambda$, 
it ends up with a simpler model with fewer number of nonzero (NNZ) 
feature weights;
for a weaker regularizer, we will get a more complex model with many more 
nonzero features weights.
From the Figure~\ref{fig:sparsity}, 
we may observe the convergence of the online learning along with the number of samples.
With a relatively larger value of~$\lambda$, the simpler model is easy to converge 
to stationary states with a small number of nonzero feature weights; 
while for a smaller~$\lambda$, we have more nonzero feature weights and
it will take many more samples for the model to reach stable states.

Figure~\ref{fig:FScoreSparsity} illustrates the trade-off between model sparsity 
and classification performance when we adjust the regularization 
parameter~$\lambda$.
For hinge loss, with a larger~$\lambda$, we get more sparse model 
at the price of a worse F-Score. 
For the log loss, as is showed in Figure~\ref{fig:FScoreSparsity}, 
it is able to prevent overfitting to some extent.
On average, it achieves the best classification performance with average F-Score 0.9174 
with the 173K (out of 1.2M) feature chosen by the sparse predictor.


\textbf {Training Errors} 


In Figure \ref{fig:errors}, we plot the number of mistakes as a function of the number of training 
samples from voted RDA and perceptron. 
The results provide empirical justifications of the theoretical analysis on error bounds described in 
Section~\ref{sec:mbd}. First, we observed that the number of training errors grows sub-linearly with the 
number of training samples. Secondly, as predicted by our analysis, voted RDA without 
regularization ($\lambda=0$) leads to less training error than perceptron, 
but would incur more errors from more regularization ($\lambda>0$).



\textbf {Single vs Average Prediction}

To investigate where the performance gain comes from, we compare the predictions of vRDA by 
single weight trained at the last sample of each epoch, and the averaged weights learned from all the training samples.
In Figure~\ref{fig:aveComp}, we plot the mean and variance bars with the corrpesonding predictions based on weights 
trained on 10 epoches. 
For both Hinge and Log losses, the average predictions have lower variance and better F-Score compared to their single 
predictions. The large variance for single predictions of Log loss implies that the predictions are quite inconsistent    
by different epoches samples; thus average predictions is highly desired here. 

\textbf {Conservative Updates}

Here, in Figure~\ref{fig:rdavrda} we compare the performance of RDA and vRDA to illustrate the trade-off 
by conservative updates with mean and variance bars based on 10 epoches.
For Hinge loss, the conservative updates (vRDA) is necessary as the hinge loss is 0 when there is classificaiton 
mistake; thus vRDA has better F-Score. While for Log loss, RDA is better as even there is a classification mistake,
we still has a non-zero loss and need to update the weights accordingly.
Another gain by conservative updates is from computational perspective. For vRDA, the frequency ratio of updating weights
is amount to the error rate of that of RDA. From our experiments, the training time 
of vRDA is about 89.7\% for Hinge loss and 87.2\% for Log loss of RDA. These precentages are not the error rate as
there are extra common computaions involved. 

\section{Conclusion and Discussions}

In this paper, we propose a voted RDA (vRDA) method to address
online classification problems with explicity regularization. 
This method updates the predictor only on the subsequence of training 
examples where a classification error is made. 
In addition to significantly reducing the computational cost involved
in updating the predictor, this allows us to derive a mistake bound
that does not depend on the total number of examples. 
We also introduce the concept of relative strength of regularization, 
and show how it affects the mistake bound and the generalization
performance.

Our analysis on mistake bound is based on the regret analysis of the RDA
method \cite{xiao:rda}.
In fact, our notion of relative strength of regularization and error bound
analysis also applies to the voted versions of other online algorithms
that admit a similar regret analysis, including the forward-backward
splitting method in \cite{DuS:09}.

We tested the voted RDA method with $\ell_1$-regularization 
on a large-scale parse reranking task in natural language processing, 
and obtained state-of-the-art classification performance with fairly sparse models.


\newpage
\clearpage
\bibliography{votedRDA}
\bibliographystyle{alpha}
\end{document}